\documentclass[runningheads]{llncs}
\usepackage{graphicx}
\usepackage{xcolor}
\begin{document}
%
\title{Domain Specific, Semi-Supervised Transfer Learning for Medical Imaging}
%
\author{Jitender Singh Virk \and Deepti R. Bathula}
%
\authorrunning{J. Virk and D. Bathula}
%
\institute{Indian Institute of Technology Ropar \\
\email{krivsj@gmail.com}\\
\email{bathula@iitrpr.ac.in}}
\maketitle              
\begin{abstract}

Limited availability of annotated medical imaging data poses a challenge for deep learning algorithms. Although transfer learning minimizes this hurdle in general, knowledge transfer across disparate domains is shown to be less effective. On the other hand, smaller architectures were found to be more compelling in learning better features. Consequently, we propose a lightweight architecture that uses mixed asymmetric kernels (MAKNet) to reduce the number of parameters significantly. Additionally, we train the proposed architecture using semi-supervised learning to provide pseudo-labels for a large medical dataset to assist with transfer learning. The proposed MAKNet provides better classification performance with $60 - 70\%$ less parameters than popular architectures. Experimental results also highlight the importance of domain-specific knowledge for effective transfer learning. 



\end{abstract}
\section{Introduction}
A large amount of annotated training data plays a critical role in making supervised deep learning models successful. For example, ResNet~\cite{resnetpaper}, a popular natural image classification architecture was trained on 1.2 million images \cite{imagenet}. When limited labeled data is available, transfer learning helps leverage knowledge from pre-trained weights as the starting point for the training process for a related task. Unfortunately, there are no large annotated medical datasets available publically as the labeling process requires domain expertise and is also costly and time-consuming. Although some public repositories exist, their small size and task-specific nature render them unsuitable for training deep architectures from scratch. 

Recent work \cite{Transfusion} not only demonstrates that transfer learning from natural image datasets to medical tasks offers limited performance gain, but also establishes that smaller architectures can learn more meaningful features and avoid over parameterization. This work attempts to address these issues with a specific focus on multi-label, body-part classification of a large medical dataset as the objective.  Our main contributions are:
\begin{itemize}
    \item We propose a novel and lightweight architecture with prudently selected modules to capture rich medical imaging features effectively.
    \item We leverage semi-supervised learning mechanism to train our architecture to help generate pseudo-labels for a relatively large medical dataset to assist with domain-specific transfer learning.
    \item We demonstrate the utility of the proposed architecture in terms of improved performance and the quality of learned features for medical imaging tasks.
\end{itemize}

\section{Dataset}
Semi-supervised learning requires both labeled as well as unlabeled data. Towards this end, we use DeepLesion \cite{deeplesiondataset} and TCIA \cite{tciasite} datasets. 

DeepLesion \cite{deeplesiondataset} is a large-scale and diverse database of lesions identified in CT scans. It has over 32k slices of dimensions $512 \times 512$. The lesions are categorized into eight types based on their coarse-scale attributes. Additionally, Yan et al. \cite{deeplesionpaper} developed a holistic approach for multi-label annotation that provides fine-grained 171 unique labels text-mined from radiology reports for this dataset. These labels are categorized into 115  body parts,  27  types,  and  29  attributes.  However, these labels are noisy and incomplete, with huge class imbalance. As this work focuses on body-part classification, we use the 115 body-part labels correspond to a total of 21511 CT images after removing noisy ones. More specifically, it has 17697 training images, 1686  validation images, and 1638 test images. Apart from the text-mined, they have a hand-labeled test set with 500 images labeled by two radiologists. After filtering only body-part related images, we acquire 490 images.

TCIA \cite{tciasite} is an extensive archive of publicly available medical datasets. Using their online \textit{Search Data Portal} along with \textit{Anatomical Site} filter, we collected CT scan volumes corresponding to 115 body part labels associated with DeepLesion dataset. Although the extracted volumes have information about the body part being examined, most of them contain slices with additional body parts making the assigned labels unusable. For example, a volume has slices from the abdomen to thigh but labeled as the bladder. Hence, we use these volumes as unlabeled data. By retaining volumes with more than 50 slices and manually removing noisy ones helped generate a little over 1.5 million unlabeled grayscale images from 6568 volumes of 3653 subjects.

\section{Mixed Asymmetric Kernels Network (MAKNet)}
\begin{figure}
\includegraphics[width=\textwidth]{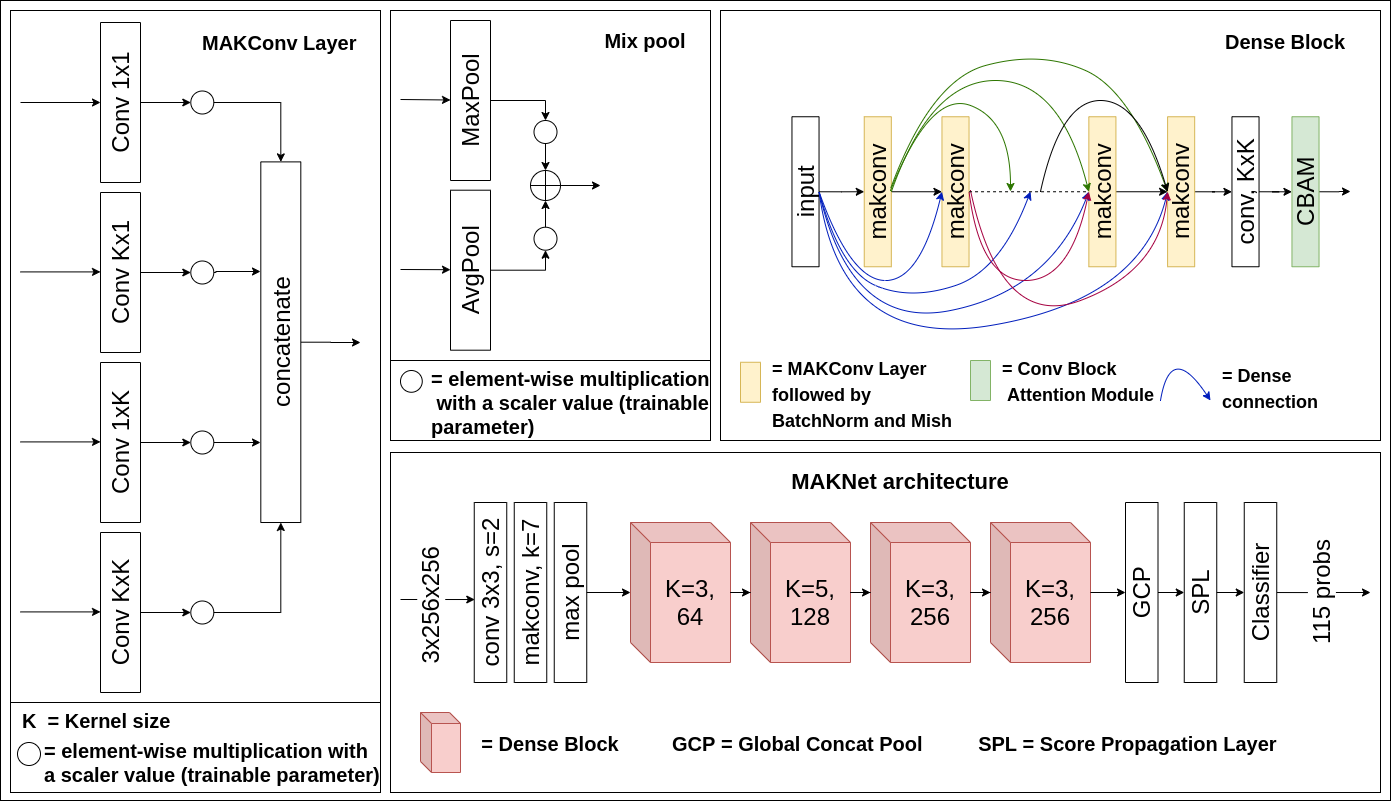}
\caption{MAKNet Architecture and its components} \label{netfig}
\end{figure}

We introduce a new type of convolution (conv) layer, which is based on mixed kernels. Similar to MixConv \cite{MixConv}, multiple kernel sizes are used in a single convolution layer. But unlike MixConv, the kernel sizes are not restricted to the form $k \times k$. Also, each kernel group processes whole input features instead of groups of them. As depicted in Figure ~\ref{netfig}, the main elements of \textit{MAKNet} are: (i) Mixed Asymmetric Kernel Convolution layer (MAKConv) - is a combination of asymmetric kernels that learns rich feature space with significantly lower number of parameters (ii) MixPool - combines max and average pooling to boost invariance to data transformations and perturbations, (iii) Global Concatenate Pooling layer (GCP)~\cite{fastaigcp} - concatenates the results of both global max and average pooling for better performance (iv) Score Propagation Layer (SPL)~\cite{deeplesionpaper} - helps enhance recall by implicitly learning the relations between labels and (v) Convolutional Block Attention Module (CBAM)~\cite{cbampaper} - a lightweight attention module for adaptive feature refinement. 

MAKConv is a combination of four asymmetric kernel types in a single conv layer. Each kernel type is divided into equal groups and targets different dimension of data. $k \times k$ target spatial features, $1 \times k$ and $k \times 1$ target single dimension aspect, and $1 \times 1$ targets depthwise features. Each kernel group accesses the entire input space. Therefore, this framework obtains a better feature space by looking at the whole input from various aspects.

Each Dense block of the MAKNet contains three or six, depending on the number of input and output features, densely connected MAKConv layers followed by a standard conv layer and CBAM attention layer as shown in Figure~\ref{netfig}: Dense Block. Each MAKConv and conv layer is followed by a BatchNorm~\cite{batchnormpaper} and Mish~\cite{mishpaper} activation function. 

\section{Experiments}
\subsection{Preprocessing}
We applied standard CT image processing steps that include converting pixel values into Hounsfield Units (HU) and normalizing them to [0,255] range. Additionally, the black-background around the scan is clipped before resizing the image to $256 \times 256$. As suggested in ~\cite{deeplesionpaper}, three consecutive slices of the volume are considered as three input channels to the network to incorporate 3D information. As no overlap is considered, this process creates a little over 500k unlabeled training images.

\subsection{Implementation Details}
Input to the network is of dimensions of $256 \times 256 \times 3$ where $3$ represents the consecutive slices as different channels. Unlike~\cite{deeplesionpaper} that uses patches around lesions, we feed the whole image as input. The total number of training images is 522476, with 17697 labeled and 504779 unlabeled. We use~\cite{deeplesionpaper}'s text-mined validation set with 1686 images and two test sets (text-mined with 1638 images and hand-labeled with 490 images) to evaluate MAKNet.

For generating pseudo-labels for unlabeled data, we retain labels from the trained teacher network with confidence $ > 0.5 $ among the top 15 predictions. Additionally, labels are filtered using exclusive relations based on the ontology information provided by~\cite{deeplesionpaper}.

All networks are trained with weighted focal loss~\cite{focallosspaper} and Ranger~\cite{rangeropt} optimizer. The teacher network is trained for 15 epochs with 0.01 learning rate. The student network is trained by concatenating labeled batch and unlabeled batch. For example, with a batch size of 64, the total images per batch would be 128. Student network is trained with 100 epochs of labeled data but approximately 4 epochs with unlabeled data to focus more on labeled data and rely less on pseudo-labels. During training, noise is introduced to the student with random rotations, grayscale, auto-contrast, and random resized cropping based data augmentation and 0.5 dropout.

\subsection{MAKNet - Multi-Label Classification Results}
To begin with, we compare the performance of our custom-designed, compact architecture that incorporates specifically selected modules for extracting rich features with popular architectures of ResNet18~\cite{resnetpaper} and VGG16~\cite{vggpaper}. The results of the comparison are shown in Table~\ref{teachertable}. As can be observed, with $\approx{70\%}$ less parameters than VGG16 and $\approx{60\%}$ less parameters than ResNet18, MAKNet provides better classification results across three of the four evaluation metrics for both text-mined and hand-labeled test sets.

\begin{table}
\centering
\caption{Multi-label classification accuracy: Results are average accuracy values across labels on two DeepLesion test sets. Bold results are the best ones.}\label{teachertable}
\resizebox{\textwidth}{!}{
\begin{tabular}{|c|c|cccc|cccc|}
\hline
\multicolumn{1}{|c|}{\textbf{Arch}} & \multicolumn{1}{|c|}{\textbf{Para-}} & \multicolumn{4}{|c|}{\textbf{Text-mined test set}} & \multicolumn{4}{|c|}{\textbf{Hand-labeled test set}}\\
\cline{3-10}
  & \textbf{meters} & AUC & F1 score & Precision & Recall & AUC & F1 score & Precision & Recall\\
\hline
VGG16        & 14.78M & 0.8912 & 0.1813 & \textbf{0.4826} & 0.4740 & 0.9030 & 0.2442 & \textbf{0.4882} & 0.5476\\
ResNet18     & 11.25M & 0.8987 & 0.1991 & 0.4285 & 0.4736 & 0.9024 & 0.2699 & 0.4503 & 0.5759\\
MAKNet &  4.50M & \textbf{0.9051} & \textbf{0.2078} & 0.4696 & \textbf{0.4744} & \textbf{0.9101} & \textbf{0.2755} & 0.4638 & \textbf{0.5780}\\
\hline
\end{tabular}
}
\end{table}

\subsection{Semi-Supervised Training Procedure}
In general, machine learning algorithms perform better if they are trained on large datasets with diverse examples. As mentioned earlier, both ResNet18 and VGG16 are trained on 1.2 million natural images. As networks trained on data from closely related domains can help extract more relevant features, we attempt to create a large medical dataset using semi-supervised learning. 

\begin{figure}
\includegraphics[width=\textwidth]{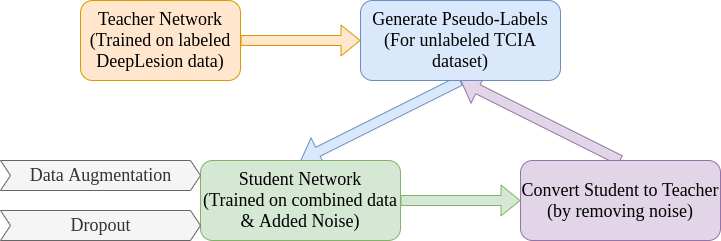}
\caption{Semi-supervised learning using noisy student method} \label{NoisyStudentFig}
\end{figure}

Noisy Student method~\cite{noisystudent} is a fairly simple but effective semi-supervised training approach. As depicted in Fig~\ref{NoisyStudentFig}, it consists of a \textit{teacher} and a \textit{student} network. First, the \textit{teacher} network is trained on a small labeled dataset and is subsequently used as a pseudo labeler to predict the labels for the large unlabeled dataset. Then, the \textit{student} network is trained on both labeled and pseudo-labeled data and iteratively tries to improve pseudo labels. ~\cite{noisystudent} suggests to add noise to the \textit{student} for better performance while training. Here, noise is simply data augmentation such as rotation, translation, cropping, etc. and dropout. Finally, we use the trained \textit{student} to pseudo-label all the unlabeled data again. Note that, during prediction, noise is disabled. The results of \textit{student} at each iteration are given in table~\ref{studenttable}.


\begin{table}
\centering
\caption{Pseudo-label iterations results of student network. The iteration 0 (teacher) represents baseline results from teacher network.}\label{studenttable}
\begin{tabular}{|c|cccc|cccc|}
\hline
\multicolumn{1}{|c|}{\textbf{Iteration}} & \multicolumn{4}{|c|}{\textbf{Text-mined test set}} & \multicolumn{4}{|c|}{\textbf{Hand-labeled test set}}\\
\hline
\multicolumn{1}{|c|}{} & AUC & F1 score & Precision & Recall & AUC & F1 score & Precision & Recall\\
\hline
0 (teacher) & 0.9051 & 0.2078 & 0.4696 & 0.4744 & 0.9101 & 0.2755 & 0.4638 & 0.5780\\
\hline
1 & 0.9074 & 0.2231 & 0.1636 & 0.6384 & 0.9153 & 0.3084 & 0.2471 & 0.7833\\
2 & 0.9065 & 0.2278 & 0.1707 & 0.6586 & 0.9177 & 0.3151 & 0.2610 & 0.7778\\
3 & 0.9093 & 0.2179 & 0.1601 & 0.6618 & 0.9181 & 0.3065 & 0.2486 & 0.7455\\
4 & 0.9064 & 0.2108 & 0.1554 & 0.7042 & 0.9164 & 0.2904 & 0.2226 & 0.8273\\
5 & 0.9112 & 0.2275 & 0.1735 & 0.6583 & 0.9102 & 0.3086 & 0.2466 & 0.7654\\
\hline
\end{tabular}
\end{table}

\subsection{Domain Specific Transfer Learning}
Transfer learning helps leverage knowledge gained from a related task by allowing the network to be initialized with pre-trained weights instead of random weights. Thus, we can get better results than training it from scratch. Moreover, pre-trained weights are precious when dealing with small datasets which is a prevalent issue in the medical domain. However, as established in \cite{Transfusion}, transferring knowledge gained from natural images is not very beneficial for medical imaging tasks.

Here we demonstrate the utility of the large pseudo-labeled dataset created using the semi-supervised technique for pre-training networks specifically for medical domain tasks. For this experiment, we consider ~\cite{deeplesionpaper}'s lesion annotation as the objective. The LesaNet developed in \cite{deeplesionpaper} uses VGG16 as the base network. Hence, we train a VGG16 on our pseudo-labeled TCIA dataset for 5 epochs. Accordingly, we initialize the base of LesaNet with our pre-trained weights and fine-tune it on DeepLesion dataset with a learning rate of 0.001 for 15 epochs. Rest of the settings for the LesaNet framework (such as ROI Pooling and Score Propagation Layers) for this experiment are the same as described in \cite{deeplesionpaper}. A comparison of performance of LesaNet with generic (natural) versus domain specific (medical) transfer learning is given in Table \ref{tltable}.


\vspace{-3mm}

\begin{table}
\centering
\caption{Comparison of transfer learning results for LesaNet pretrained on ImageNet and TCIA. Both are fine-tuned on DeepLesion dataset. 
}\label{tltable}
\resizebox{\textwidth}{!}{%
\begin{tabular}{|l|cccc|cccc|}
\hline
\multicolumn{1}{|c|}{\textbf{Method}} & \multicolumn{4}{|c|}{\textbf{Text-mined test set}} & \multicolumn{4}{|c|}{\textbf{Hand-labeled test set}}\\
\cline{2-9}
\multicolumn{1}{|c|}{} & AUC & F1 score & Precision & Recall & AUC & F1 score & Precision & Recall\\
\hline
LesaNet (ImageNet) & \textbf{0.9344} & 0.3423 & \textbf{0.3593} & 0.5327 & 0.9398 & 0.4344 & 0.4737 & 0.5274\\
LesaNet (TCIA) & 0.9331 & \textbf{0.3569} & 0.3045 & \textbf{0.6669} & \textbf{0.9403} & \textbf{0.4972} & \textbf{0.4754} & \textbf{0.7531}\\
\hline
\end{tabular}
}
\end{table}

\vspace{-3mm}

\section{Understanding Network Predictions}
Interpretability of a model is a crucial factor for acceptance in sensitive domains like healthcare. Consequently, in addition to the quantitative results above, we attempt to explain the significance of the network’s decisions as follows.

\vspace{-3mm}

\subsection{Integrated Gradients}

\begin{figure}
\centering
\includegraphics[width=0.85\textwidth]{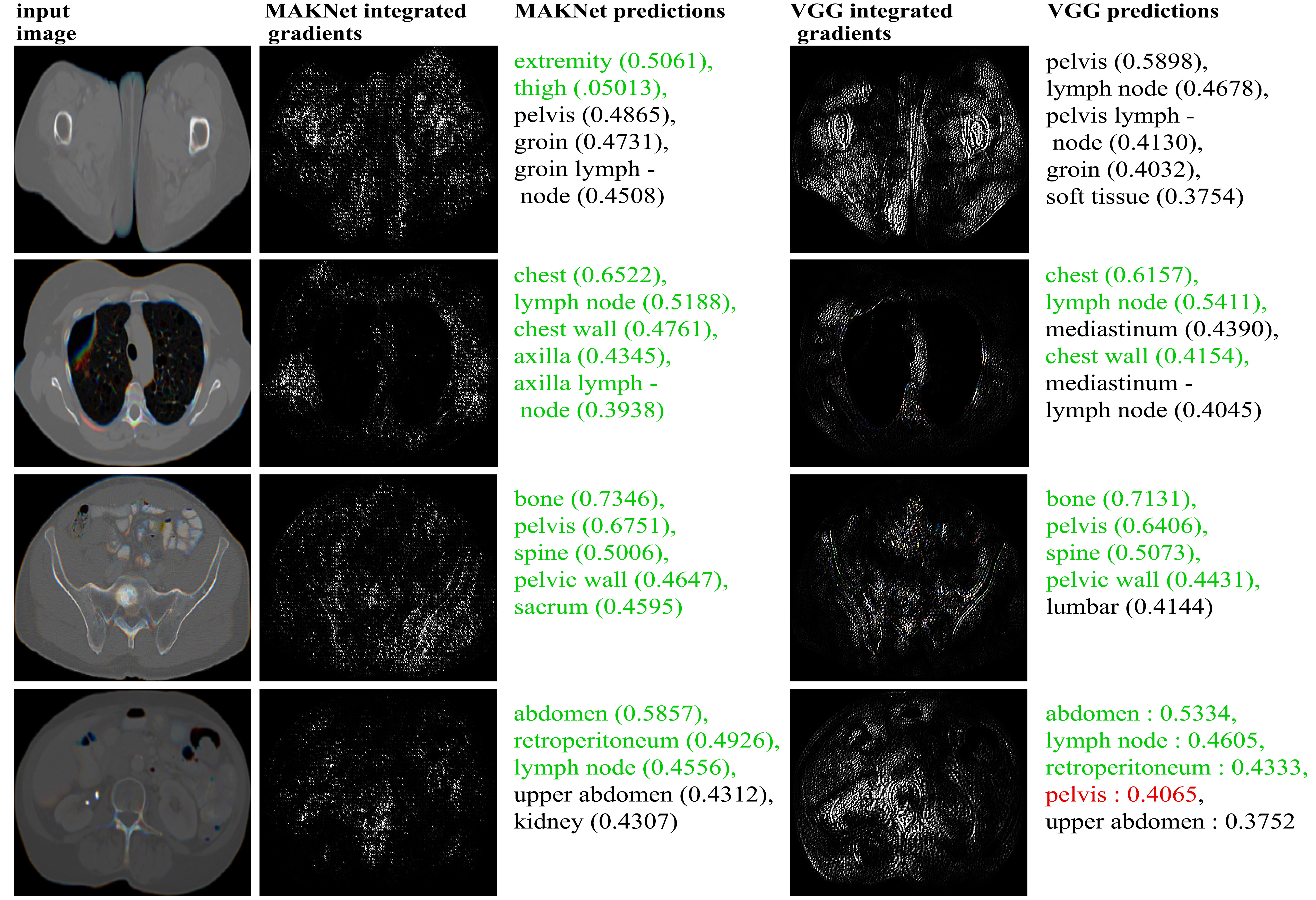}
\caption{Examples are showing the integrated gradients with top 5 predicted labels and corresponding confidences for MAKNet (column 2) and VGG16 (column 3). Green and Red correspond to True Positive (TP) and False Positive (FP) labels, respectively. Black represents labels with missing annotations as not all images in the dataset are assigned 5 labels. Gradients are scaled for visualization. Figure best visible in colors.} \label{igfig}
\end{figure}

Gradient-based attribution method has been used often to quantify feature importance in linear models. However, with deep and nonlinear models, it suffers from saturation, causing the important features to have tiny gradients. Integrated Gradients~\cite{integradpaper} is a recently introduced alternative that avoids saturation. By creating a sequence of linearly interpolated images from baseline (black) to the input image and integrating the gradients of the output with respect to these of series of interpolated images, this method provides both sensitivity to changes as well as invariance to implementation. In this work, we used 50 steps for interpolation. A qualitative comparison of integrated gradients for MAKNet and VGG16, along with their predictions, is depicted in Figure ~\ref{igfig}.


\subsection{Perturbation}
\begin{figure}
\includegraphics[width=\textwidth]{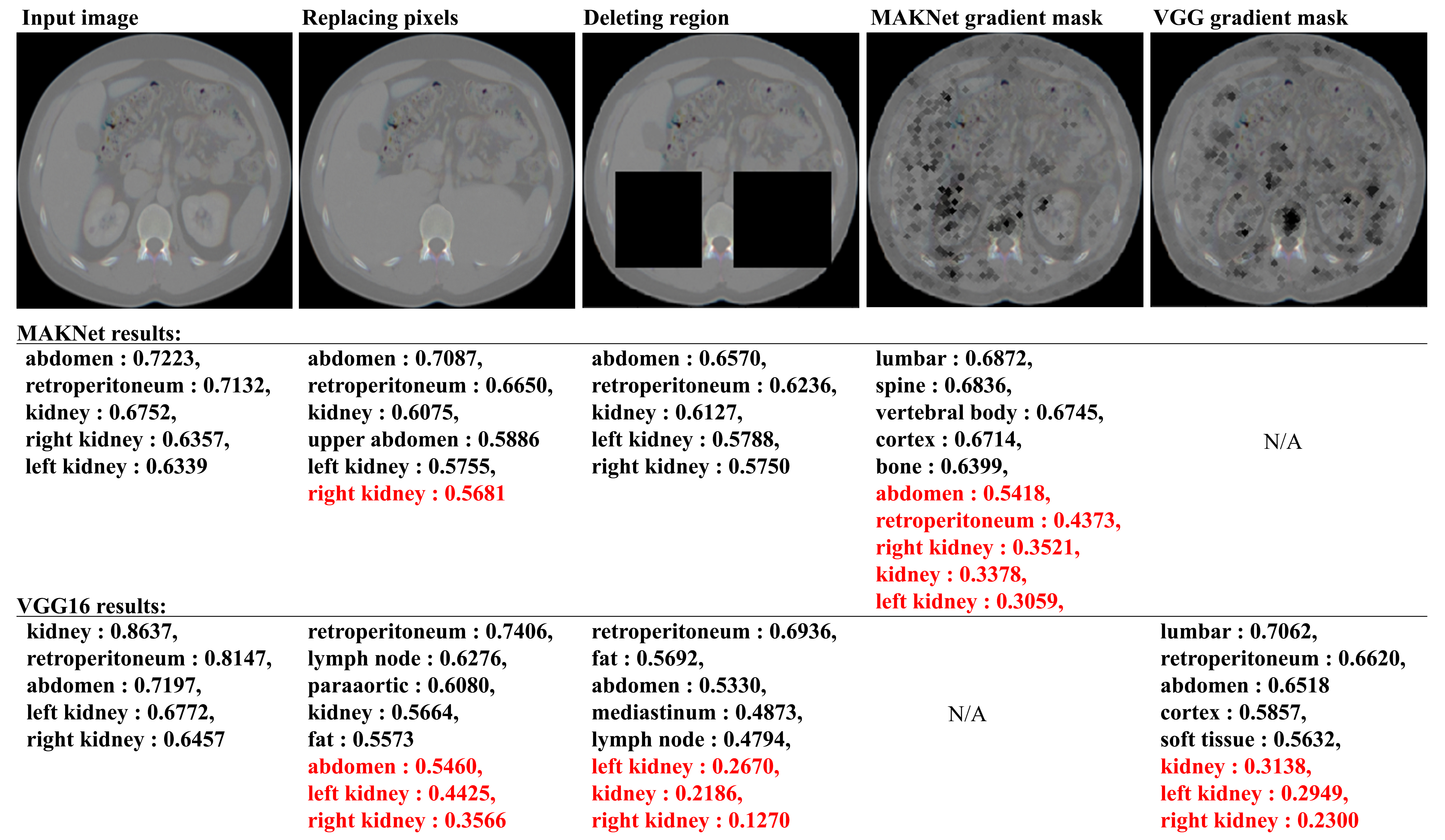}
\caption{Perturbation Example: Top row shows perturbed images. Middle and bottom rows show top 5 predictions (with confidence) made by MAKNet and VGG16 respectively. Red color identifies labels whose confidence is affected by perturbation, causing them to drop below the top 5.
} \label{ptbfig}
\end{figure}

Another complementary attribution method relies on measuring the effect of perturbations applied to the input image. In this work, we use three different perturbation strategies: (a) replacing an individual organ's pixels with neighbour pixels, (b) applying black patches in regions of interest, and (c) deleting most influential pixels with a mask generated by integrated gradients. Figure~\ref{ptbfig} shows one example of a perturbed image with corresponding top 5 predictions and confidence levels. Relative to VGG16, as MAKNet learns the implicit relationship between labels, no significant change in label confidence level is observed for replacement and deletion. In contrast, masking salient regions based on integrated gradients caused a significant drop in the confidence of the original top 5 labels highlighting the importance of the features learned by the MAKNet.

\section{Conclusion and Future Work}
In conclusion, we demonstrate the utility of our proposed lightweight architecture with $60 - 70\%$ less parameters in providing improved multi-label, body-part classification accuracy. We have also interrogated the network using attribution and perturbation techniques to understand its predictions. Using semi-supervised learning, we trained our network to provide pseudo-labels for a large unlabeled medical dataset and showed its effectiveness for transfer learning in the medical domain. Going forward, we hope to create a domain-specific pre-trained network that offers improved performance across various medical imaging modalities and tasks.



%
%

\begin{thebibliography}{8}

\bibitem{resnetpaper}
He, K., Zhang, X., Ren, S., Sun, J.: Deep Residual Learning for Image Recognition. CoRR. abs/1512.03385 (2015) \url{http://arxiv.org/abs/1512.03385}

\bibitem{vggpaper}
Simonyan, K., Zisserman, A.: Very Deep Convolutional Networks for Large-Scale Image Recognition. International Conference on Learning Representations, 2015 \url{https://www.robots.ox.ac.uk/~vgg/publications/2015/Simonyan15/simonyan15.pdf}

\bibitem{imagenet}
Deng, J., Dong, W., Socher, R., Li, L.-J., Li, K., Fei-Fei, L.: ImageNet: A Large-Scale Hierarchical Image Database. IEEE Computer Vision and Pattern Recognition (CVPR), 2009 \url{http://www.image-net.org/}

\bibitem{deeplesiondataset}
Yan, K., Wang, X., Lu, L., Summers, R.: DeepLesion: automated mining of large-scale lesion annotations and universal lesion detection with deep learning. Journal of Medical Imaging \textbf{5}(3), 1--11 (2018) \doi{10.1117/1.JMI.5.3.036501}

\bibitem{deeplesionpaper}
Yan, K., Peng, Y., Sandfort, V., Bagheri, M., Lu, Z., Summers, R.: Holistic and Comprehensive Annotation of Clinically Significant Findings on Diverse {CT} Images: Learning from Radiology Reports and Label Ontology. CoRR. abs/1904.04661 (2019) \url{http://arxiv.org/abs/1904.04661}

\bibitem{Transfusion}
Raghu, M., Zhang, C., Kleinberg, J., Bengio, S.: 
Transfusion: Understanding Transfer Learning with Applications to Medical Imaging, CoRR. abs/1902.07208 (2019) \url{http://arxiv.org/abs/1902.07208}

\bibitem{MixConv}
M. Tan and Quoc V. Le: MixConv: Mixed Depthwise Convolutional Kernels. CoRR. abs/1907.09595 (2019) \url{http://arxiv.org/abs/1907.09595}

\bibitem{mishpaper}
Misra, D.: Mish: A Self Regularized Non-Monotonic Neural Activation Function. CoRR. 2019 \url{arXiv:1908.08681v2}

\bibitem{batchnormpaper}
Ioffe, S., Szegedy, C.: Batch Normalization: Accelerating Deep Network Training by Reducing Internal Covariate Shift. ICML. \textbf{37}, 448-456 (2015) \url{http://proceedings.mlr.press/v37/ioffe15.html}

\bibitem{tciasite}
Clark, K., Vendt, B., et al: The Cancer Imaging Archive (TCIA): Maintaining and Operating a Public Information Repository. Journal of Digital Imaging. 26, 1045–1057 (2013) \url{https://doi.org/10.1007/s10278-013-9622-7}

\bibitem{noisystudent}
Xie, Q., Hovy, E., Luong, M.T., Le, Q.: Self-training with Noisy Student improves ImageNet classification. ArXiv. abs/1911.04252 (2019)
\url{arXiv:1911.04252}

\bibitem{focallosspaper}
Lin, T.Y., Goyal, P., Girshick, R., He, K., Doll, P.: Focal Loss for Dense Object Detection. CoRR. abs/1708.02002 (2017) \url{http://arxiv.org/abs/1708.02002}

\bibitem{integradpaper}
Sundararajan, M., Taly, A., Yan, Q.: Axiomatic Attribution for Deep Networks. CoRR. abs/1703.01365 (2017) \url{http://arxiv.org/abs/1703.01365}

\bibitem{cbampaper}
Woo, S., Park, J., Lee, J.Y., Kweon, I.: CBAM: Convolutional Block Attention Module. CoRR. abs/1807.06521 (2018) \url{http://arxiv.org/abs/1807.06521}

\bibitem{rangeropt} Ranger Optimizer Homepage, \url{https://tinyurl.com/wp6ve3f}. Last accessed 10 Jan 2020

\bibitem{fastaigcp}
Adaptive Concat Pool Homepage, \url{https://tinyurl.com/vmulq75}. Last accessed 27 Jan 2020






\end{thebibliography}
%

\end{document}